\documentclass[letterpaper]{article} 
\usepackage{aaai24}  
\usepackage{times}  
\usepackage{helvet}  
\usepackage{courier}  
\usepackage[hyphens]{url}  
\usepackage{graphicx} 
\usepackage{natbib}  
\usepackage{caption} 
\usepackage{algorithm}

\usepackage{array,multirow}
\usepackage{amsmath}
\usepackage{mathtools}
\usepackage{hhline}
\usepackage{boldline}
\usepackage{tabularx}
\usepackage{comment}
\usepackage{amssymb}
\usepackage{booktabs}
\usepackage{xcolor}   
\usepackage{boldline}
\usepackage{amssymb}
\usepackage{algorithm}
\usepackage{algpseudocode}
\usepackage{soul} 
\usepackage[capitalize]{cleveref}
\usepackage{subcaption}
\usepackage{kotex}
\usepackage{caption}
\usepackage{makecell}
\newcolumntype{P}[1]{>{\centering\arraybackslash}p{#1}}
\newcolumntype{L}[1]{>{\arraybackslash}p{#1}}
\usepackage{times}

\usepackage{newfloat}
\usepackage{listings}
\DeclareCaptionStyle{ruled}{labelfont=normalfont,labelsep=colon,strut=off} 
\floatstyle{ruled}
\newfloat{listing}{tb}{lst}{}
\floatname{listing}{Listing}
\title{Task-Disruptive Background Suppression for Few-Shot Segmentation}
\author{
    Suho Park, SuBeen Lee, Sangeek Hyun, Hyun Seok Seong, Jae-Pil Heo\thanks{Corresponding author}\\
}
\affiliations{
    Sungkyunkwan University\\
    \{shms0706, leesb7426, hsi1032, gustjrdl95, jaepilheo\}@skku.edu
}

\begin{document}

\maketitle

\begin{abstract}
Few-shot segmentation aims to accurately segment novel target objects within query images using only a limited number of annotated support images. 
The recent works exploit support background as well as its foreground to precisely compute the dense correlations between query and support.
However, they overlook the characteristics of the background that generally contains various types of objects.
In this paper, we highlight this characteristic of background which can bring problematic cases as follows:
(1) when the query and support backgrounds are dissimilar and (2) when objects in the support background are similar to the target object in the query.
Without any consideration of the above cases, adopting the entire support background leads to a misprediction of the query foreground as background.
To address this issue, we propose Task-disruptive Background Suppression~(TBS), a module to suppress those disruptive support background features based on two spatial-wise scores: query-relevant and target-relevant scores.
The former aims to mitigate the impact of unshared features solely existing in the support background, while the latter aims to reduce the influence of target-similar support background features.
Based on these two scores, we define a query background relevant score that captures the similarity between the backgrounds of the query and the support, and utilize it to scale support background features to adaptively restrict the impact of disruptive support backgrounds.
Our proposed method achieves state-of-the-art performance on PASCAL-5$^i$ and COCO-20$^i$ datasets on 1-shot segmentation.
Our official code is available at github.com/SuhoPark0706/TBSNet.
\end{abstract}
\section{Introduction}
\label{Sec.1 Intro}
\begin{figure}[t!]
    \centering
    \includegraphics[width=\columnwidth]{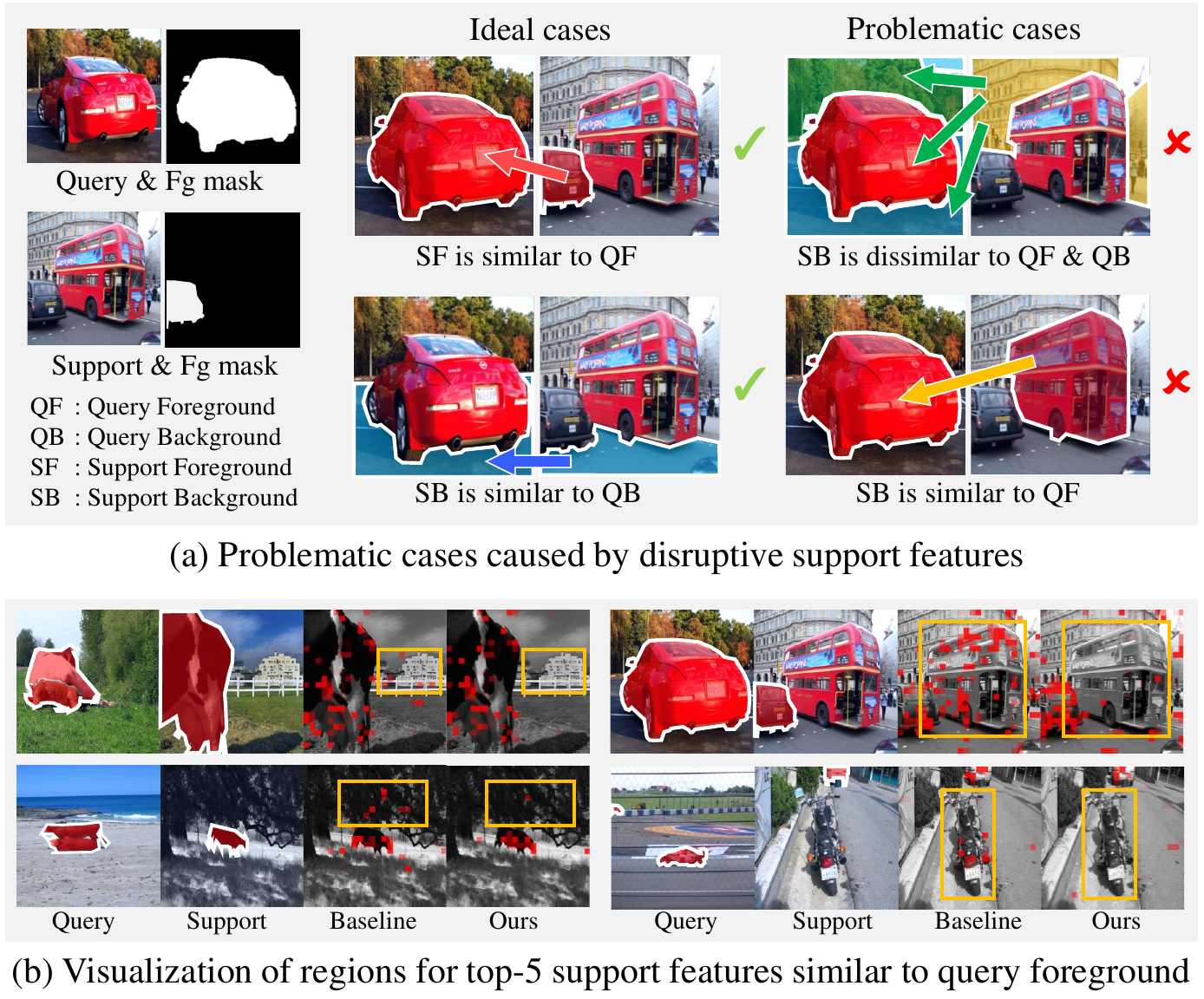}
    \caption{
    The abbreviation Q, S, F, B are query, support, foreground and background, respectively.
    (a) Two types of disruptive SB can occur: 
    (1) SB dissimilar to all Q can introduce ambiguity in the relationship between QB and SB.
    (2) SB more similar to QF than QB can cause misclassification of QF as background.
    (b) We visualize the top-5 support regions similar to QF at the feature level, the closer to the top-1 the darker.
    In the first column, we illustrate cases with query-irrelevant SB, while the second column demonstrates scenarios where SB highly resembles QF.
    In both cases, our approach effectively suppresses disruptive SB that exhibits high similarity to QF, as indicated by the yellow boxes.
    A more detailed analysis is shown in \cref{sec:averaged attention}.
    }
    \label{fig:motivation}
\end{figure}

\begin{figure}[t!]
    \centering
    \includegraphics[width=\columnwidth]{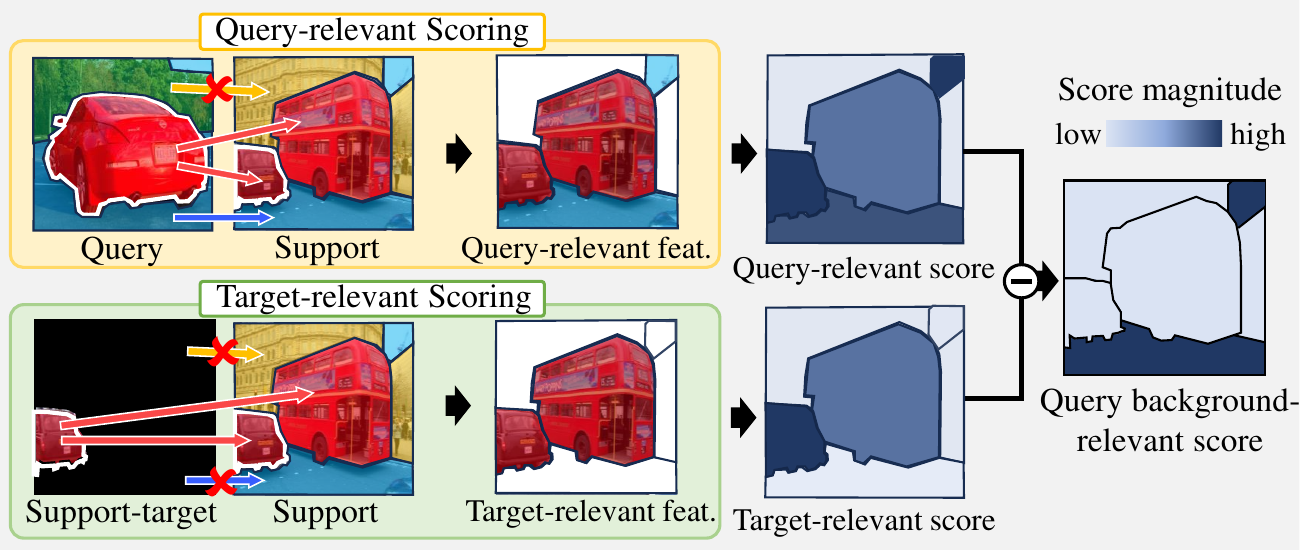}
    \caption{
    Illustration of two types of scoring processes. Query-relevant scoring aims to reduce the influence of unshared features solely existing in the support background~(yellow region) and target-relevant scoring is to restrict the influence of target-similar support background features~(red bus).
    Query background-relevant score is computed based on these two scores to focus on shared background features~(blue and skyblue regions in support image).
    }
    \label{fig:spatial-wise score representation}
\end{figure}

With the advance of deep learning, semantic segmentation~\cite{seg0, seg1,seg2, HP} has achieved remarkable performance.
However, its performance relies on abundant data for target classes and degrades noticeably with insufficient data. 
To resolve this, Few-Shot Segmentation~(FSS) has been proposed~\cite{shaban2017one} to build a model adaptable for novel classes with only a few number of labeled data.
Briefly, FSS aims to learn a novel class with a small number of labeled images, called a support set, to segment an unlabeled image, called a query set.

In Few-Shot Segmentation, affinity learning is a mainstream technique~\cite{min2021hypercorrelation_hsnet, shi2022dense_DCAMA, zhang2021few_cyctr, DAN}, learning pixel-wise correlations. 
Early methods~\cite{DAN, min2021hypercorrelation_hsnet} utilize only the support foreground~(SF) features to compute the correlations with the query features~(Q).
However, they overlook that the support background~(SB) features also contain contextual information helpful for distinguishing between the query foreground~(QF) and background~(QB).
For instance, the sky can be particularly useful to segment airplanes.
Therefore, the most recent techniques~\cite{zhang2021few_cyctr, shi2022dense_DCAMA} leverage the entire set of support features~(S), encompassing both foreground~(SF) and background~(SB).

However, it is important to note that not every SB feature is beneficial for distinguishing between the QF and QB.
Since a background can contain diverse objects, the dissimilarity between the QB and SB can disrupt the segmentation. 
Moreover, the SB may include objects similar to the target object in the query image. 
In such cases, the high similarity between the SB and QF may mislead the model to predict QF pixels as QB.
Therefore, we need to filter out these harmful SB pixels.

In this context, we propose Task-disruptive Background Suppression (TBS), a module designed to suppress harmful background features within a support set to enhance the segmentation of the query.
Specifically, to determine the utility of background features, TBS defines two spatial-wise scores: query-relevant and target-relevant scores. 
The query-relevant score determines the similarity between each SB feature and the entire Q features, calculated using the cross-attention module~\cite{CrossTransformer}.
This score allows us to identify SB pixels relevant to the Q image while filtering out irrelevant ones.
However, SB features similar to the QF may still persist, potentially leading to the misprediction of the QF mask.
To address it, we introduce the target-relevant score for each SB feature, indicating the degree of similarity with the target object features in the support.
Similar to the query-relevant score, the target-relevant score is derived through the cross-attention between each SB feature and SF feature.
By combining these two scores, we define a query background-relevant score; the relevance score of SB features to QB.
As a result, we suppress task-disruptive SB features that are irrelevant to a given QF or similar to the target object class. This is achieved by multiplying these scores with SB features.
To sum up, our contributions are summarized as follows:
\begin{itemize}
    \item For the first time, we define the advantageousness of support background features based on the relation between background regions of query and support images. 
    \item We propose a novel module,  Task-disruptive Background Suppression (TBS) that restricts certain background features within the support set that do not contribute to precise segmentation of the query image.
    \item Our method achieves state-of-the-art performance over baselines, and its effectiveness is validated by various ablation studies.
\end{itemize} 

\section{Related Work}
\subsection{Few-Shot Segmentation}
The methods of Few-Shot Segmentation~(FSS) can be categorized into two main streams: prototype-based and affinity learning methods.
Prototype-based methods~\cite{ASGNet, zhang2020sg-one,wang2019panet,liu2020ppnet} represent the foreground objects in the support set as single or multiple prototypes.
They classify the pixels of the query image into foreground and background based on their similarity to the prototypes.
However, these methods may lead to deteriorated segmentation results, since they lose information about support objects while summarizing the images with only a few representative features. 
On the other hand, affinity learning methods~\cite{min2021hypercorrelation_hsnet,hongcost_VAT,wang2023rethinking_abcnet} leverage pixel-level dense correlations between the object features of support set and query features.
Moreover, recent techniques~\cite{zhang2021few_cyctr, shi2022dense_DCAMA} have found that the background features of the support set are also useful in distinguishing between the foreground and background of the query.
Therefore, they compute correlations with query features using not only the object features but also the background features.
Although they show impressive performance by exploiting whole support features, they still overlook that some background features are not useful for classifying the foreground and background of the query image.
On the other hand, our method assigns low weights to those task-disruptive background features to prevent the pixel of the query image from being misclassified. 

\begin{figure*}[t]
    \centering
    \includegraphics[width=0.97\linewidth]{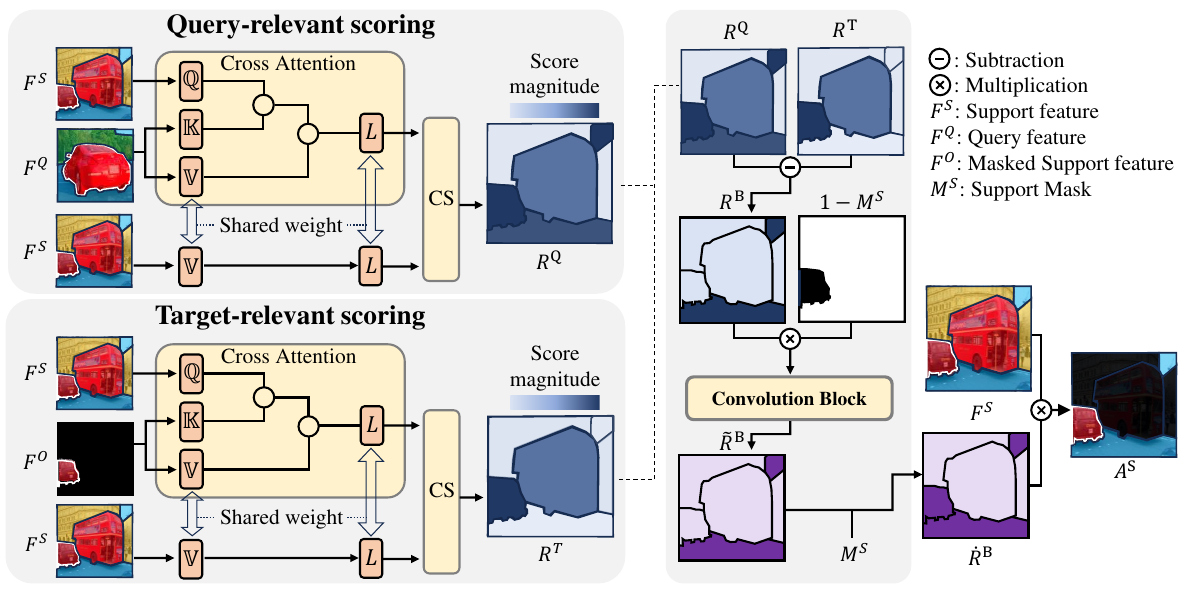}
    \caption{Illustration of Task-disruptive Background Suppression~(TBS) module. TBS incorporates two scoring processes to compute query-relevant and target-relevant scores. 
    By representing support features $F^S$ based on query features $F^Q$ by the cross-attention, we can get the query-relevant scores $R^\text{Q}$ which signify how well each support feature describes the query feature. On the other hand, the target-relevant score $R^\text{T}$ assesses whether each support background feature is similar to the target object feature, obtained by representing the support feature with the support object feature. By employing the score refinement by $R^\text{Q}-R^\text{T}$, query background relevant score is computed to activate regions that are query-relevant but target-irrelevant. Consequently, by applying $\dot{R}^\text{B}$ to the original support features $F^S$ through multiplication, we can obtain adaptive support features $A^S$.}
    \label{fig:overall_architecture}
\end{figure*}

\subsection{Feature Suppression in FSS}
The cross-attention module in few-shot segmentation is generally employed to constrain the disruptive support features in distinguishing between the query foreground and background~\cite{wang2023rethinking_abcnet, hdmnet, zhang2021few_cyctr}.
For example, CyCTR~\cite{zhang2021few_cyctr} utilizes bidirectional cross-attention to identify the most similar support feature for each query feature and vice versa. 
Consequently, when the category of a support feature (the starting feature) differs from the class of the most similar support feature (the ending feature) for the corresponding query feature, the starting feature is identified as potentially disruptive and is mitigated.
similary, ABCNet~\cite{wang2023rethinking_abcnet} suppresses unuseful support features by indirectly comparing the support and query features through computing cross-attention with a reference pool.
However, these methods may struggle to demonstrate effectiveness especially when the support background is similar to the query foreground. This limitation arises because they do not consider the categorically derived relationship between features.
Although our method also restrains disruptive support features like existing methods based on cross-attention, we utilize the relationship between the support foreground and background for the first time.
By incorporating this novel relationship in addition to the relation between the support and query, we can discover task-disruptive support features with consideration of only the similarity between the support background and the query background.

\section{Our Method}
The overall architecture of our method is illustrated in \cref{fig:overall_architecture}.
Affinity learning methods in FSS require precise pairing between the query and support features, especially based on their binary class~(foreground and background). However, as noted in the Introduction, backgrounds of different images typically consist of various objects and may not share common characteristics, in contrast to the foregrounds.
Thus, in this paper, we aim to refine support background features by considering the following two conditions.
First, the support background features similar to the query background features are preferred. 
Another is that the support background features should be distinguished from the query object features.
In other words, these two conditions intend to reduce the gap between backgrounds of support and query, and enhance the disparity between background and foreground simultaneously, encouraging the query and support features to be well-clustered according to their binary class.

To meet these requirements, we introduce two representativeness scores: query-relevant score $R^\text{Q}$ and target-relevant score $R^\text{T}$, which are pixel-level importance scores for support background features.
Specifically, $R^\text{Q}$ signifies whether the support background features can describe the query features. 
Thus, the support background features well represented by the query features would have high $R^\text{Q}$.
However, since the query features used in computing $R^\text{Q}$ contain both the query background and foreground, it cannot be guaranteed that $R^\text{Q}$ is derived only from the background.
To resolve this, we define $R^\text{T}$ which is highly activated when the support background features are similar to the target object features in the support set.
By subtracting $R^\text{T}$ from $R^\text{Q}$, we filter out the scores activated by the query target object at $R^\text{Q}$. 
We define the filtered scores as background-relevant score $R^\text{B}$ which restrains task-disruptive support background features containing the information of the target object.
We utilize $R^\text{B}$ to selectively suppress the influence of the support background features.    

In the following sections, we provide the problem definition and a detailed explanation of the aforementioned scores for refining the support background features.

\subsection{Problem Definition}
As a standard formulation of the few-shot segmentation problem, we have two disjoint datasets: $D_{\text{train}}$ for training a model and $D_{\text{test}}$ for evaluating a learned model. Each dataset consists of distinct object classes $C_{\text{train}}$ and $C_{\text{test}}$ without any overlap $\left(C_{\text{train}} \cap C_{\text{test}}=\emptyset\right)$.
Generally, the training and testing of few-shot segmentation are composed of several episodes.
Each episode consists of $K$ labeled images and an unlabeled image, i.e., $K$-shot episode, where all images contain objects of the same category $c$ randomly sampled from the dataset.
Specifically, the labeled images are called a support set $S=\left\{\left(I^S_j,M^S_j\right)\right\}^{K}_{j=1}$, and the unlabeled image is named a query set $Q=\left(I^Q,M^Q\right)$. 
Here, $I$ is an image and $M$ denotes a corresponding ground-truth binary mask, containing a value of 1 for foreground object regions, belonging to category $c$, and 0 for the others.
The goal of the few-shot segmentation is to predict $M^Q$ based on $S$.

\subsection{Spatial-wise Representativeness Scores}
For features of the $j$-th support image $F^S_j$ computed by a feature extractor, we spatially divide $F^S_j$ into support object features $F^{O}_j$ and support background features $F^{B}_j$ by ground-truth mask $M^S_j$, as follows:
\begin{equation}
    \begin{split}
        \label{eq:spt_fg_bg}
        F^O_j&=\left\{f^S_{j,h,w}|m^S_{j,h,w}=1\right\}
        \\
        F^B_j&=\left\{f^S_{j,h,w}|m^S_{j,h,w}\ne1\right\},
    \end{split}
\end{equation}
where $f^S_{j,h,w}$ and $m^S_{j,h,w}$ denote the value spatially located at $\left(h,w\right)$ of $F^S_j$ and $M^S_j$, respectively.
Among these two sorts of features, $F^O_j$ can be treated as valuable features for the task, since they only contains the categorical information of target objects that we need to detect in the query set $Q$. In contrast, $F^B_j$ may include both useful and unuseful features, so thus it is crucial to prevent the unuseful features within $F^B_j$ from being utilized in the segmentation process. Therefore, we introduce a query-relevant score $R^\text{Q}$ of how much each pixel well represents query-relevant information.

To measure the relevance of each pixel, we utilize the cross-attention module~\cite{CrossTransformer}.
However, unlike the existing cross-attention, we reconstruct the support features $F^S_j$ based on the query features $F^Q$.
Specifically, reconstructed support features $\bar{F}^S_j$ are computed as follows:
\begin{equation}
   \begin{split}
        \label{eq:recon}
        \bar{F}^S_j=\mathbb{V}\left(F^Q\right)
        \text{Softmax}\left(\frac{\mathbb{Q}\left(F^S_j\right)
        {\mathbb{K}\left(F^Q\right)}^{T}}{\sqrt{d}}\right),
    \end{split}
\end{equation}
where $d$ is the channel dimension of projection, and $\mathbb{Q}$, $\mathbb{K}$, and $\mathbb{V}$ are linear heads for query, key, and value, respectively.

Since the reconstruction quality of $F^S_j$ would be high when the support feature is similar to the query features, we define the similarity between $\bar{F}^S_j$ and $\mathbb{V}\left(F^S_j\right)$ with additional linear projection as the ${R}^\text{Q}_j$ which represents query-relevance score of $j$-th support feature $F^S_j$, as follows:
\begin{equation}
   \begin{split}
        \label{eq:cs}
        {R}^\text{Q}_j=\text{CS}\left(L(\bar{F}^S_j), L({\mathbb{V}\left(F^S_j\right)})\right),
    \end{split}
\end{equation}
where $\text{CS}\left(\cdot,\cdot\right)$ is the cosine-similarity function, and $L$ is a linear projection layer.

Although we can estimate how much each pixel of support features is relevant to query features based on ${R}^\text{Q}$, it is not solely derived from its similarity with the query background.
In this case, the query-relevance score alone can be problematic, since the high similarity between the support background and query object induces the pixels of the query objects to be predicted as the background.
To resolve this issue, we define a complementary score, a target-relevant score ${R}^\text{T}$, which indicates whether each pixel is similar to the target object.
We first compute reconstructed support features $\hat{F}^S_j$ with the support object features $F^O_j$, as follows:
\begin{equation}
   \begin{split}
        \label{eq:recon_2}
        \hat{F}^S_j=\mathbb{V}\left(F^O_j\right)
        \text{Softmax}\left(\frac{\mathbb{Q}\left(F^S_j\right)
        {\mathbb{K}\left(F^O_j\right)}^{T}}{\sqrt{d}}\right).
    \end{split}
\end{equation}
Then, $R^\text{T}_j$ is determined the similarity between $\hat{F}^S_j$ and $\mathbb{V}\left(F^O_j\right)$, as follows:
\begin{equation}
   \begin{split}
        \label{eq:cs_2}
        R^\text{T}_j=\text{CS}\left(L(\hat{F}^S_j), L({\mathbb{V}\left(F^S_j\right)})\right).
    \end{split}
\end{equation}

Note that, the parameters of projection heads such as $L$, $\mathbb{Q}$, $\mathbb{K}$, and $\mathbb{V}$ are shared with Eq.~\ref{eq:recon}

\begingroup
\begin{table*}[t]
    \centering
    \begin{tabular}{l | c | c c c c c c }
        \hlineB{2.5}
        \multirow{2}{*}{\textbf{Method}} & 
        \multirow{2}{*}{\textbf{Backbone}} & 
        \multicolumn{6}{c}{\textbf{1-shot/5-shot}}
        \\
        & & \textbf{F-0} & \textbf{F-1} & \textbf{F-2} & \textbf{F-3} & \textbf{mIoU} & \textbf{FB-IoU}
        \\
        \hlineB{2.5}
        CWT ~\cite{CWT} \ & \multirow{8}{*}{ResNet101} & 56.9/62.6 & 65.2/70.2 & 61.2/68.8 & 48.8/57.2 & 58.0/64.7 & - \\
        DoG-LSTM ~\cite{DoG-LSTM} \ & & 57.0/57.3 & 67.2/68.5 & 56.1/61.5  & 54.3/56.3 & 58.7/60.9 & - \\
        DAN ~\cite{DAN} \ & & 54.7/57.9 & 68.6/69.0 & 57.8/60.1 & 51.6/54.9 & 58.2/60.5 & 71.9/72.3 \\
        HSNet ~\shortcite{min2021hypercorrelation_hsnet} \ & & 67.3/71.8 & 72.3/74.4 & 62.0/67.0 & 63.1/68.3 & 66.2/70.4 & 77.6/80.6 \\
        \cline{0-0} \cline{3-8} 
        CyCTR~\cite{zhang2021few_cyctr} \ & & 67.2/71.0 & 71.1/75.0 & 57.6/58.5 & 59.0/65.0 & 63.7/67.4 & - \\
        CyCTR + TBS (Ours) \ & & 67.8/72.3 & 70.9/74.7 & 57.8/59.5 & 59.6/65.5 & 64.0/68.0 & - \\
        \cline{0-0} \cline{3-8} 
        DCAMA~\cite{shi2022dense_DCAMA} \ & &65.4/70.7 & 71.4/73.7 & 63.2/66.8 & 58.3/61.9 & 64.6/68.3 & 77.6/80.8 \\
        DCAMA + TBS (Ours) \ & & 68.5/72.3 & 72.0/74.1 & 63.8/68.4 & 59.5/67.2 & 65.9/70.5 & 77.7/81.3 \\     
        \hlineB{2.5}
        HSNet ~\shortcite{min2021hypercorrelation_hsnet} \ & \multirow{3}{*}{Swin-B}  &67.9/72.2 & 74.0/\textbf{77.5} & 60.3/64.0 & 67.0/72.6 & 67.3/71.6 & 77.9/81.2 \\
        DCAMA~\cite{shi2022dense_DCAMA}\ & &72.2/75.7 & 73.8/77.1&64.3/\textbf{72.0}&67.1/74.8&69.3/74.9&78.5/82.9\\
        DCAMA + TBS (Ours) \ & &\textbf{74.7}/\textbf{76.5}&\textbf{74.4}/76.5&\textbf{66.1}/71.4&\textbf{69.5}/\textbf{76.3}&\textbf{71.2}/\textbf{75.2}&\textbf{80.0}/\textbf{83.4}\\
        \hlineB{2.5}
    \end{tabular}
    \caption{
    Experimental results on the PASCAL-5$^i$ dataset with ResNet-101 and Swin-B backbones.}
    \label{tab:pascal}
\end{table*}
\endgroup

\begingroup
\begin{table*}[t]
    \centering
    \begin{tabular}{l | c | c c c c c c }
        \hlineB{2.5}
        \multirow{2}{*}{\textbf{Method}} & 
        \multirow{2}{*}{\textbf{Backbone}} & 
        \multicolumn{6}{c}{\textbf{1-shot/5-shot}}
        \\
        & & \textbf{F-0} & \textbf{F-1} & \textbf{F-2} & \textbf{F-3} & \textbf{mIoU} & \textbf{FB-IoU}
        \\
        \hlineB{2.5}
        HSNet ~\shortcite{min2021hypercorrelation_hsnet} \ & \multirow{3}{*}{Swin-B} &43.6/50.1&49.9/58.6&49.4/56.7&46.4/55.1&47.3/55.1&72.5/76.1\\
        DCAMA~\cite{shi2022dense_DCAMA} & &49.5/\textbf{55.4} &52.7/60.3 &52.8/\textbf{59.9} &48.7/57.5 &50.9/58.3 &73.2/\textbf{76.9} \\
        DCAMA + TBS (Ours) \ & &\textbf{49.6}/53.7 &\textbf{54.3}/\textbf{62.9} &\textbf{54.1}/59.3 &\textbf{51.3}/\textbf{58.2} &\textbf{52.3}/\textbf{58.5} &\textbf{74.2}/76.8\\
        \hlineB{2.5}
    \end{tabular}
    \caption{
    Experimental results on the COCO-20$^i$ dataset with Swin-B backbone.
    }
    \label{tab:coco}
\end{table*}
\endgroup
\subsection{Task-disruptive Background Suppression}
We first define a query background-relevant score $R^\text{B}$ through the subtraction of $R^\text{T}$ from $R^\text{Q}$ and multiply it with a support background mask $1 - M^S$ element-wisely.
Since the subtraction operation signifies the removal of target object relevance from query relevance, the score $R^\text{B}$ reflects the similarity of each pixel in the support background features to the query background features. 
We convert this score into spatial-wise weights to suppress disruptive support background features as follows:
\begin{equation}
   \begin{split}    
        \label{eq:bs}
        \tilde{R}^\text{B}_j=b\left(R^\text{B}_j\right),
    \end{split}
\end{equation}
where $b$ is a shallow convolution block for refinement and the architecture of it is explained in \cref{Sec:implementation_details}.

Recall that our goal is to refine the support background features to resolve an undesirable background matching problem. 
However, directly multiplying the score map $\tilde{R}^\text{B}_j$ with the support feature $F^S_j$ would result in the incomplete preservation of foreground features.
To prevent this, we replace the value of $\tilde{R}^\text{B}_j$ corresponding to $M^O_j$ with 1, denoting it as $\dot{R}^\text{B}_j$.
Consequently, we obtain adaptive support features $A^S_j$ by multiplying $\dot{R}^\text{B}_j$ with $F^{S}_j$.
We then utilize $A^S_j$ in the subsequent segmentation process instead of $F^S_j$.

\section{Experiments}
\subsection{Datasets and Evaluation Metrics}
\begin{figure*}
    \centering
    \includegraphics[width=1.\linewidth]{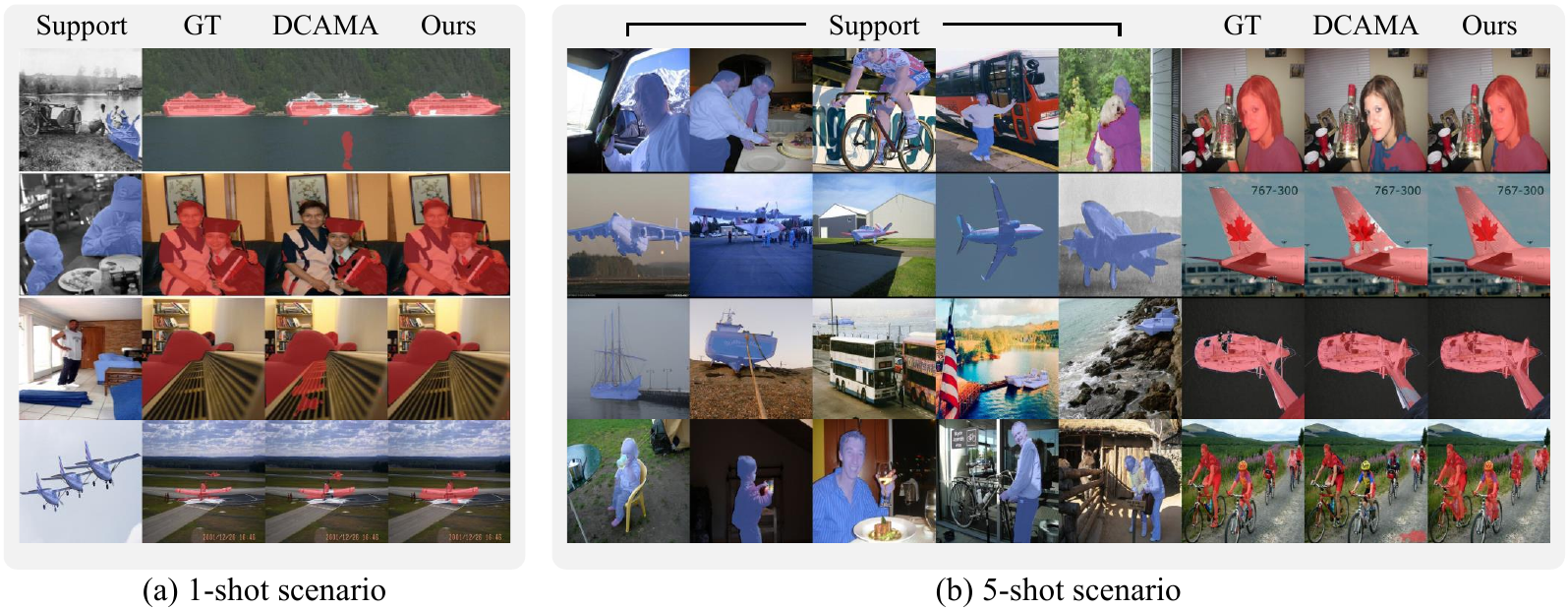}
    \caption{
    Qualitative comparison results of DCAMA and DCAMA + TBS (Ours) on PASCAL-5$^i$ dataset with Swin-B.
    }
    \label{qualitative}
\end{figure*}

We utilize PASCAL-5$^i$~\cite{shaban2017one-shot_fss} and COCO-20$^i$\cite{cocofewshot} following the prior works~\cite{zhang2021few_cyctr, shi2022dense_DCAMA, hdmnet, wang2023rethinking_abcnet}.
PASCAL-5$^i$ combines data from PASCAL VOC 2012~\cite{pascal} and SDS~\cite{sds}, comprising 20 categories.
In contrast, COCO-20$^i$ is a subset of COCO~\cite{coco} and is comprised of 80 categories.
And, each dataset is divided into four folds where it has the same number of categories that do not overlap with others.
Hence, each fold of PASCAL-5$^i$ and COCO-20$^i$ have 5 and 20 classes, respectively.
To evaluate the model's adaptability to novel classes, we adopt a cross-validation scheme where each fold is selected as $D_{\text{test}}$ and others are used as $D_{\text{train}}$.
Then, we evaluate the model with mean intersection over union~(mIoU) and foreground-background intersection over union~(FB-IoU) for 1000 episodes randomly sampled from $D_{\text{test}}$.

\subsection{Implementation Details}
\label{Sec:implementation_details}

To verify the high adaptability of TBS, we apply it to two baseline models: DCAMA~\cite{shi2022dense_DCAMA} and CyCTR~\cite{zhang2021few_cyctr}.
For fair comparisons with baselines, we adopt ResNet-101 pretrained on ImageNet and Swin-Transformer pre-trained on ImageNet 1K as a feature extractor.
In the case of DCAMA with Swin-Transformer, we apply TBS at scales of $\frac{1}{8}$, $\frac{1}{16}$, and $\frac{1}{32}$ to align with the scale used in DCAMA's cross-attention mechanism.
However, for DCAMA with ResNet-101, we utilize TBS only in $\frac{1}{16}$ and $\frac{1}{32}$ scales due to memory limitation.
On the other hand, since CyCTR was verified only on ResNet, we conducted experiments on ResNet-101, not Swin-Transformer.
Unlike DCAMA which adopts multi-level features, CyCTR utilizes single-level features generated by combining features from 3- and 4-th blocks.
Therefore, we suppress only those combined features by using TBS.
Many hyper-parameters, \textit{i.e.}, optimizer, learning rate, batch size, etc., are the same as the baseline.

We describe the flow of the convolution block for converting the query background-relevant score into the spatial-wise weight. 
It first concatenates the input score map and the layer normalization of the score map to reference the spatial-wise distribution of scores.
Subsequently, two consecutive layers of $1\times1$ convolution without any non-linear activation are employed. This projects the input score maps into 256 channels and squeezes it into 1 channel again. After that, a sigmoid function is applied to make the score map into the range between 0 and 1. 

\subsection{Experimental Results}

\paragraph{Quantitative Results.}
We evaluate our proposed method by comparing it with previous techniques designed for few-shot segmentation. 
As illustrated in \cref{tab:pascal}, recent affinity learning models, specifically CyCTR and DCAMA, already exhibit comparable performances. 
Upon incorporating TBS into these approaches, a consistent improvement over baseline model performances is observed, resulting in the state-of-the-art scores. 
This improvement remains consistent across various evaluation metrics and different quantities of labeled images on the PASCAL-5$^i$ dataset. 

Similar trends are observed in the 1-shot scenario of COCO-20$^i$. 
As demonstrated in \cref{tab:coco}, TBS consistently enhances DCAMA's performance across all folds, providing the best performance. 
While its impact is less pronounced in the 5-shot scenario compared to the 1-shot scenario, where it shows substantial effectiveness, TBS still succeeds in improving the average mIoU of DCAMA.

As a result, TBS surpasses the existing state-of-the-art performance in three out of four quantitative benchmark scenarios in the context of few-shot segmentation.
This verifies the effectiveness of suppressing disruptive support, particularly in situations of extreme data scarcity.

\paragraph{Qualitative Results.}
In addition to the quantitative results, we report qualitative results to intuitively show the effectiveness of TBS.
Compared with DCAMA, our results include fewer mispredicted pixels regardless of the number of support images as shown in \cref{qualitative}.
Especially, when objects in the support background are not present in the query background, our model outperforms DCAMA.
This validates that our method appropriately suppresses unnecessary support background. Additional in-depth analysis of it is provided in \cref{subsec:scoreMap_visualization}.

\begin{figure*}
    \centering
    \includegraphics[width=1.0\linewidth]{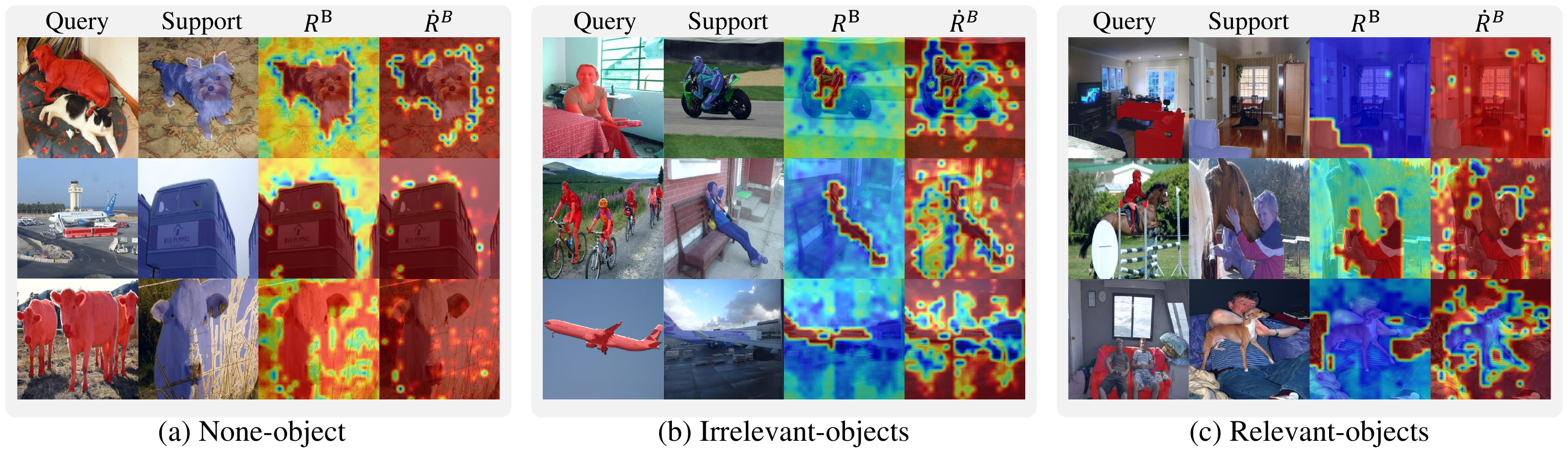}
    \caption{
    Visualization of the background-relevant scores before and after the score refinement module.
    The color from red to blue represents high to low scores. $R^\text{B}$ denotes score maps before refinement, while $\dot{R}^\text{B}$ denotes score maps after refinement.
    }
    \label{fig:BS_visualization}
\end{figure*}


\begin{table}[ht]
  \begin{subtable}{0.42\linewidth}
        \centering
        \begin{tabular}{c c c}
            \hlineB{2.5}
            QS & TS & mIoU 
            \\
            \hlineB{2.5}
            - & - & 73.8 \\
             \checkmark & - & 73.4 \\
            - & \checkmark & 74.1 \\
            \checkmark & \checkmark & \textbf{74.4} \\
            \hlineB{2.5}
        \end{tabular}
    	\caption{Effect of two spatial-wise scores. QS and TS denote the query- and target-relevant scores.}
    	\label{ablation study}
  \end{subtable}%
  \hspace{0.01\linewidth}
  \begin{subtable}{0.56\linewidth}
    \centering
    \begin{tabular}{l | c | c}
        \hlineB{2.5}
        Method & Metric & AA 
        \\
        \hlineB{2.5}  
        \multirow{3}{*}{Baseline} & SF\&QF & 0.213 
        \\
         & SB\&QB & \textbf{0.886}
         \\
         & Avg. & 0.550
         \\
         \hline
         \multirow{3}{*}{Ours} & SF\&QF & \textbf{0.325}
         \\
         & SB\&QB & 0.820 
         \\
         & Avg. & \textbf{0.573}
         \\
         \hline
        \hlineB{2.5}
    \end{tabular}
    \caption{Averaged Attention score (AA)~(See \cref{sec:averaged attention}).}
    \label{fig: attention aggregation analysis}
  \end{subtable}
  \caption{Results of ablation studies}
\end{table}

\section{Further Analysis}
In this section, we conduct ablation studies and provide an in-depth analysis of our method. 
For most ablation studies, we use the PASCAL-5$^i$ dataset in the 1-shot scenario with Swin-B as the backbone network, except for \cref{kshot}.
Additionally, mIoU is adopted for metric which is one of the most standard metrics in few-shot segmentation.

\begin{figure}
    \centering
        \centering
        \includegraphics[width=\columnwidth]{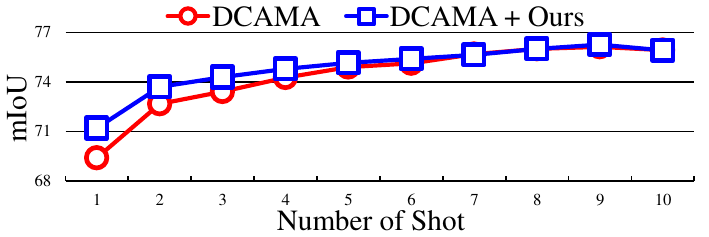}
        \caption{
        $K$-shot results with varying $K$. 
        }
        \label{fig: kshot}
\end{figure}

\subsection{Ablation Study on Main Components}
\cref{ablation study} presents the effects of query- and target-relevant scores in fold-1. 
In the second row, the query-relevant score reduces mIoU by 0.4\%. 
We argue that the query-relevant score can inherently possess a negative effect by emphasizing some of the support backgrounds similar to the target object.
Therefore, suppressing the support background with only the query-relevant score might be harmful to performance.
Thus, the query-relevant score should be used in conjunction with the target-relevant score. 
On the contrary, as shown in the third row, the target-relevant score enhances the baseline by 0.3\%. 
It means that some support background similar to the target object resembles the query foreground, making feature suppression advantageous. 
Importantly, adopting both scores can further boost performance, affirming their complementary nature.

\subsection{Varying K for K-shot}
\label{kshot}
We validated the merits of our method using standard evaluation protocol with 1 and 5 labeled images (i.e., $K=1$ and $5$).
We also conducted experiments with varying numbers of labeled images and the results are provided in \cref{fig: kshot}.
As reported, the benefits of our method are especially highlighted at low shots in terms of relative performance improvements. 
It confirms that our method is notably more effective in scenarios with severe data scarcity, aligning well with the requirements of few-shot learning.

\subsection{Quality of Feature Matching} 
\label{sec:averaged attention}

Following our motivation, TBS suppresses the task-disruptive features, enhancing the similarity between query and support foreground~(QF and SF), and also between query and support background~(QB and SB).
To analyze these changes in similarity caused by TBS, we examine the cross-attention map in the segmentation model~(DCAMA).

Specifically, we average the attention scores corresponding to the QF \& SF pairs to capture their similarity, and perform the same computation for the QB \& SB pairs as well.
As shown in \cref{fig: attention aggregation analysis}, we observe that the proposed method achieves a higher averaged attention score compared to the baseline.
This observation implies that ours successfully improves the similarity between foreground objects.
However, for the attention score of background pairs, ours achieve lower score compared to baseline.
We suspect that unintended suppression of useful background occurs, resulting in lower attention scores of background pairs.
Nevertheless, we verify the improvement when averaging the scores of both pairs and highlight the significant enhancement in attention scores of foreground pairs.

\subsection{Visualization of Background-relevant Score}
\label{subsec:scoreMap_visualization}
We visualize background-relevant scores under diverse conditions to verify TBS in mitigating task-disruptive support background regions, as shown in \cref{fig:BS_visualization}.
In the first scenario where objects do not exist in support background, as demonstrated in \cref{fig:BS_visualization}~(a), TBS only restrains the support object boundaries that are treated as background.
The next scenario is when objects within the support background are not present in the query background.
As shown in \cref{fig:BS_visualization}~(b), TBS assigns low scores to these regions, demonstrating the impact of query-relevant scoring.
In the last scenario, shared objects are present in both the support and query backgrounds.
These shared objects act as helpful features since they enhance the similarity between query and support backgrounds. As depicted in \cref{fig:BS_visualization}~(c), we verify that TBS grants high scores to leverage these features for segmentation.
More importantly, TBS exhibits its effectiveness even in scenarios where relevant and irrelevant objects coexist within the support background~(last row of \cref{fig:BS_visualization}~(c)).
In such a case, we expect that the segmentation model may exploit only the relevant objects while suppressing the irrelevant ones to enhance background similarity.
As shown in the last row of \cref{fig:BS_visualization}~(c), although the support backgrounds contain both people and a dog, only the dog is suppressed, as people are present in the query image while the dog is not.
To sum up, TBS effectively suppresses task-disruptive support backgrounds in various conditions.

Additionally, we compare background-relevant scores before and after convolution block to analyze the effectiveness of the refinement module.
Notably, scores before refinement tend to have small variations between background-relevant and -irrelevant objects, potentially hindering effective suppression due to a subtle difference in suppression power.
In contrast, the refined score maps present a wider spectrum of magnitudes, decisively influencing the determination of whether pixels in the support background warrant suppression.
Furthermore, without the refinement module, the score distribution across the query-support image pairs exhibits significant diversity.
This disparity is evident when comparing the background score distributions from \cref{fig:BS_visualization}~(a) and (c).
Conversely, the distribution of scores after refinement attains consistency across the query-support pairs, enhancing their stability as input for the segmentation model.

\section{Conclusion}
In this paper, we introduce a Task-Disruptive Background Suppression module designed to mitigate the problem that query-irrelevant and target-similar features in support background regions. 
To suppress such features, we present two types of scores. 
First, a query-relevant score is employed to filter out irrelevant support background pixels whose similarity between query images is low. 
Second, a target-relevant score is used for detecting support backgrounds that are similar to the support foreground.
Based on these two score maps, we could suppress task-disruptive backgrounds in the support set. Finally, experiments conducted on standard benchmarks show the effectiveness of our model.

\section*{Acknowledgments}
This work was supported in part by MSIT/IITP (No. 2022-0-00680, 2019-0-00421, 2020-0-01821, 2021-0-02068), and MSIT\&KNPA/KIPoT (Police Lab 2.0, No. 210121M06).

\bibliography{Park}

\end{document}